%% file: main.tex
\documentclass[10pt,twocolumn,letterpaper]{article}

\usepackage{cvpr}              
\usepackage{comment}
\usepackage[export]{adjustbox}
\input{preamble}

\definecolor{cvprblue}{rgb}{0.21,0.49,0.74}
\usepackage[pagebackref,breaklinks,colorlinks,allcolors=cvprblue]{hyperref}
\usepackage{mdframed}
\usepackage{xcolor}
\usepackage{multirow}

\def\confName{CVPR}
\def\confYear{2026}

\title{LiteEmbed: Adapting CLIP to Rare Classes}

\author{
    \textit{Aishwarya Agarwal}$^{1,2}$\thanks{\scriptsize \textit{aishwarya.agarwal@research.iiit.ac.in}, \textit{aishagar@adobe.com}} \quad
    \textit{Srikrishna Karanam}$^{2}$\thanks{\scriptsize \textit{skaranam@adobe.com}} \quad
    \textit{Vineet Gandhi}$^{1}$\thanks{\scriptsize \textit{vgandhi@iiit.ac.in}}\\[2mm]
    \small $^{1}$CVIT, Kohli Centre for Intelligent Systems, IIIT Hyderabad, India\\
    \small $^{2}$Adobe Research, Bengaluru, India
}

\begin{document}


\twocolumn[{
\renewcommand\twocolumn[1][]{#1}%
\maketitle
\begin{center}
 \centering
 \captionsetup{type=figure}
 \includegraphics[width=0.96\textwidth]{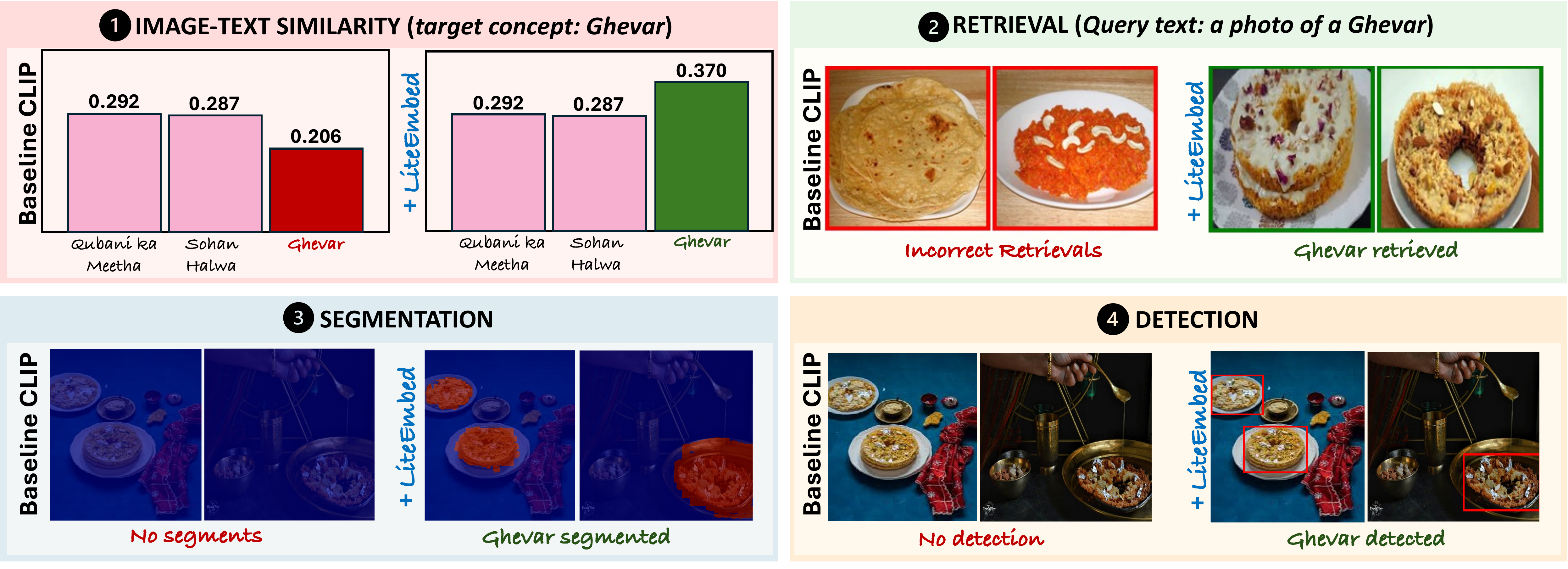}
 \caption{Given a target concept, we introduce~\modelname, which refines CLIP’s text embedding through Subspace-Guided Optimization using only 3–5 reference images \includegraphics[height=1.5em,valign=c]{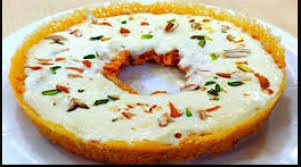} \includegraphics[height=1.5em,valign=c]{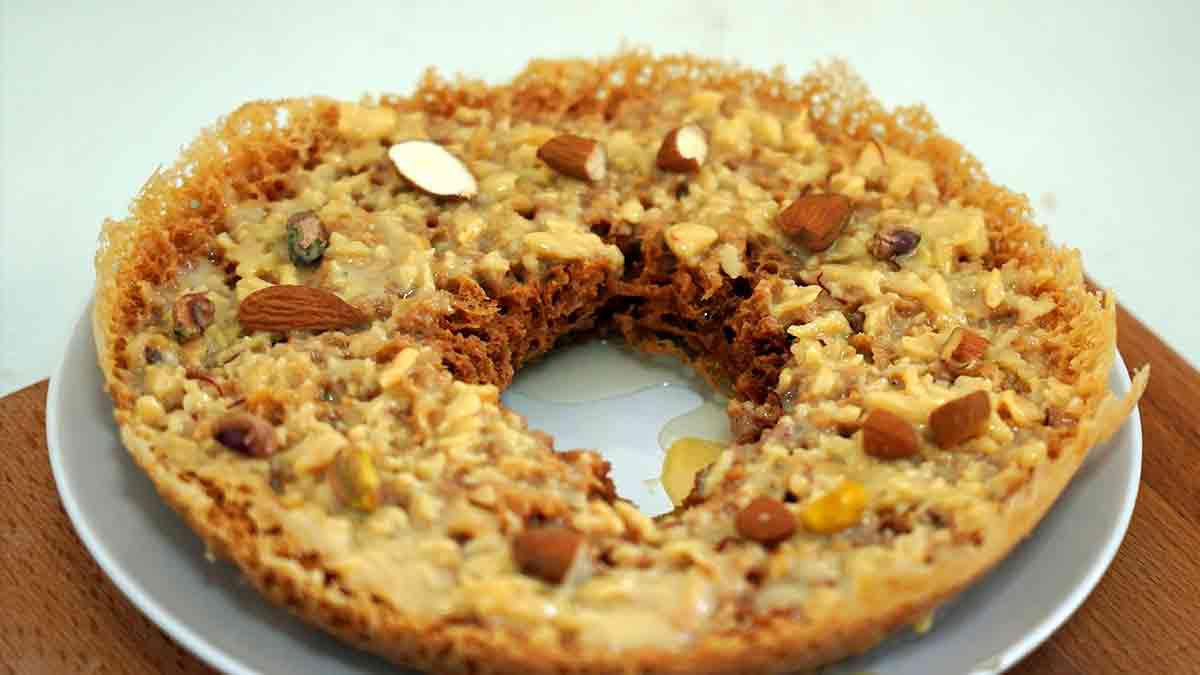} \includegraphics[height=1.5em,valign=c]{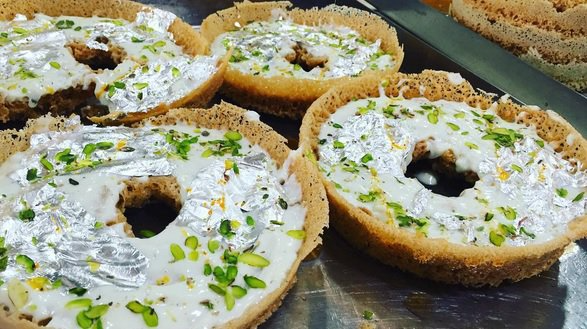}. The optimized embedding plugs directly into CLIP-based models, here applied to classification, retrieval, segmentation~\cite{luddecke2022image}, and detection~\cite{cheng2024yolo}, and consistently improves performance without task-specific retraining.  
}
 \label{fig:teaser}
\end{center}
}]


\input{sec/0_abstract}    
\input{sec/1_intro}
\input{sec/2_relatedWorks}
\input{sec/3_methodology}
\input{sec/4_results}
\input{sec/5_conclusion}


{
    \small
    \bibliographystyle{ieeenat_fullname}
    \bibliography{longstrings, main}
}

\clearpage

\input{sec/X_suppl}

\end{document}

%% file: preamble.tex
\newcommand{\tabpub}[1]{{\scriptsize \color{gray}{#1}}}

\newcommand{\modelname}{LiteEmbed}



%% file: sec/0_abstract.tex
\begin{abstract}
Large-scale vision–language models such as CLIP achieve strong zero-shot recognition but struggle with classes rarely seen in pretraining, including newly emerging entities and culturally specific classes. We introduce \modelname, a lightweight framework for few-shot personalization of CLIP that enables new classes to be added without retraining its encoders.~\modelname~ performs subspace-guided optimization of text embeddings within CLIP’s vocabulary, leveraging a PCA-based decomposition that disentangles coarse semantic directions from fine-grained variations. Two complementary objectives, coarse alignment and fine separation, jointly preserve global semantic consistency while enhancing discriminability among visually similar classes. Once optimized, the embeddings are plug-and-play, seamlessly substituting CLIP’s original text features across classification, retrieval, segmentation, and detection. Extensive experiments demonstrate substantial gains over prior methods, establishing \modelname~ as an effective approach for adapting CLIP to underrepresented, rare or unseen classes.
\end{abstract}

%% file: sec/1_intro.tex
\section{Introduction}
\label{sec:intro}
\begin{figure}[t]
\hfill
\begin{center}
\includegraphics[width = 1.01\linewidth]{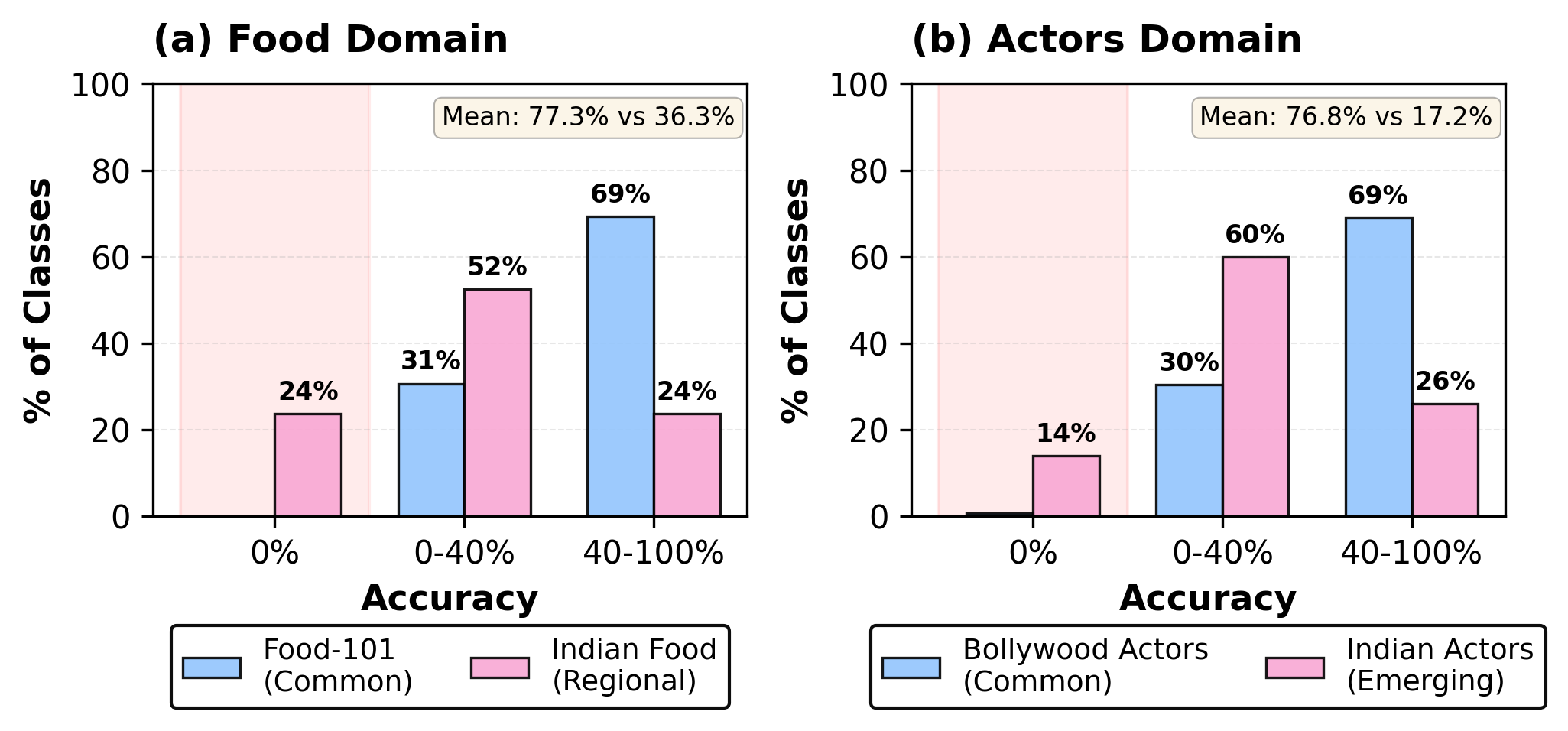}
\end{center}
\caption{Performance disparity in CLIP’s zero-shot classification.}
\label{fig:introFig}
\end{figure}

Large-scale Vision--Language Models (VLMs) like CLIP~\cite{radford2021learning} have brought about an era of powerful zero-shot visual recognition. Yet, their knowledge is limited to training domains: newly emerging classes (e.g., a recently prominent actor) or culturally specific items (e.g., regional Indian dishes) remain poorly represented.~\cref{fig:introFig} reports CLIP’s zero-shot classification accuracy, showing that while common Food-101~\cite{bossard2014food} and Indian Actor classes~\cite{indian-actor-images} score high, 24\% of regional dishes and 14\% of emerging actors receive zero recognition, with most others below 40\%. This highlights systematic failure on underrepresented classes and a critical gap in real-world use-cases.

To bridge this gap, we explore extending CLIP to new classes using only 3–5 images, without retraining its large encoders or relying on prior data. Our goal is to make new classes immediately usable across any task or pipeline that leverages CLIP’s text embeddings, from classification and retrieval to generation, segmentation, and detection. Achieving this requires preserving CLIP’s core strengths: (i) alignment, maintaining a coherent cross-modal representation; (ii) discrimination, preserving clear separability among visually similar fine-grained classes; and (iii) compositionality, enabling flexible use of new classes within natural prompts such as ``a boy eating \texttt{Ghevar}". In essence, the adapted model should operate as if these classes had been part of CLIP’s pretraining, generalizing seamlessly while retaining full downstream utility.

Prior efforts to adapt CLIP have shown strong task-specific gains but are not directly suited for few-shot, extensible class addition. Adapter-based approaches~\cite{gao2024clip, zhang2022tip} insert small trainable modules into CLIP’s frozen backbone to specialize on tasks such as classification. While efficient, these modules modify the shared representation space, making the adapted features task-specific and less generalizable to other applications such as retrieval, scoring, or generation. Prompt tuning methods~\cite{zhou2022conditional, zhou2022learning, khattak2023maple, feng2023diverse, yoon2024c, abdul2023align} learn contextual tokens to improve alignment across known classes but rely on large-scale supervision and shared context embeddings rather than class-specific ones, limiting discrimination and compositionality in few-shot, single-class settings. Moreover, these methods assume static training distributions and cannot extend CLIP incrementally. Together, these limitations highlight the need for approaches that can flexibly introduce new classes into CLIP’s embedding space, while preserving its core properties of alignment, discrimination, and compositionality.

Building on these insights, we introduce \textbf{\modelname}, a lightweight optimization framework for adapting CLIP to rare classes. Each new class is represented by a learnable token within CLIP’s vocabulary, individually optimized using few-shot examples. We observe that directly aligning a new embedding with the few-shot examples by maximizing similarity often leads to two key failures: (a) the embedding may diverge from its semantic neighborhood (e.g., \texttt{Golden Retriever} separating from other dog-related classes), or (b) collapse toward visually similar categories (e.g., \texttt{Labrador Retriever}), thereby reducing fine-grained discriminability. We posit that semantic alignment and discrimination operate at distinct scales: alignment should preserve coarse global semantics (e.g., recognizing a \texttt{Golden Retriever} as a dog, not a cat), while discrimination should capture subtle distinctions among related breeds. Iterestingly, a PCA analysis of CLIP’s text embeddings supports this hypothesis, revealing that high-variance directions encode coarse categorical structure, whereas low-variance directions capture fine-grained variations (see~\cref{subsec:pca}).

To this end,~\modelname{} employs subspace-guided optimization, which leverages CLIP’s semantic structure through two complementary loss terms. For each novel class (e.g., \texttt{Golden Retriever}), we assemble a semantic neighborhood of related coarse classes (e.g., \texttt{dog, canine}) and a visually confusable set of fine-grained negatives (e.g., \texttt{Labrador Retriever, Irish Setter}). PCA on their combined text embeddings defines a local subspace that disentangles coarse, high-variance directions from fine-grained, low-variance ones. We then introduce (i) a \textbf{Coarse Alignment Loss} that aligns the embedding with the semantic neighborhood along high-variance components, and (ii) a \textbf{Fine Separation Loss} that enforces separation from confusable classes along low-variance components. These subspace-aware objectives yield a lightweight, training-free adaptation of CLIP to new classes, requiring only minimal optimization over few-shot examples at test time.

\modelname{} produces embeddings that are fully plug-and-play: once optimized, they can seamlessly substitute CLIP’s original text embeddings across diverse downstream tasks. This includes classification, retrieval, segmentation, and detection, all without task-specific retraining.~\cref{fig:teaser} demonstrates substantial gains using \modelname{} in retrieving accurate images, generating faithful masks, detecting correct objects, and producing compositional generations. 

We comprehensively evaluate \modelname{} across classification and retrieval tasks on three publicly available datasets: TV100~\cite{zhou2024tv100}, Korean Celebrities~\cite{seo2023new}, and Indian Food Images~\cite{food2022indian}. To facilitate a more diverse and rigorous evaluation of adaptation to novel and underrepresented classes, we further introduce \textbf{NOVA} (New Object and Visual Alignment), a benchmark comprising five meticulously curated datasets. NOVA encompasses game characters, emerging landmarks, fashion ensembles, and recently prominent public figures, thereby establishing a realistic and challenging testbed for few-shot generalization. In addition, \modelname{} is assessed on UECFood100~\cite{bossard2014food} for segmentation and detection tasks, and on CustomConcept101~\cite{kumari2023multi} for text-to-image generation. Formally, our key contributions are as follows:

\begin{itemize}
    \item We propose \modelname{}, a Subspace-Guided Optimization framework that adapts CLIP to rare classes from a few visual exemplars. PCA is employed to disentangle CLIP’s text-embedding manifold into global and local semantic subspaces, within which novel-class token embeddings are optimized through two complementary losses that jointly preserve coarse alignment and enhance fine-grained discriminability.
    \item We present comprehensive quantitative, qualitative, and ablative experiments across ten datasets and five tasks, demonstrating that \modelname{} enables plug-and-play use of learned embeddings, directly substituting CLIP text features without retraining and yields consistent, pronounced gains over existing methods.
\end{itemize}

%% file: sec/2_relatedWorks.tex
\section{Related Works}
\label{sec:related_works}

Large-scale vision–language models~\cite{radford2021learning,jia2021scaling,li2022blip,li2023blip,xiao2024florence,tschannen2025siglip} have established strong cross-modal alignment but remain limited to their pretraining domains. A large body of work has explored adapting CLIP to new distributions through prompt or adapter tuning. Prompt-based methods~\cite{zhou2022conditional,zhou2022learning,khattak2023maple,lu2022prompt,zhu2023prompt} learn context tokens to better align textual and visual features, while adapter-based methods~\cite{gao2024clip,zhang2022tip} insert lightweight modules for efficient fine-tuning. Despite strong performance, these approaches depend on multiclass supervision and often modify CLIP’s shared embedding space, limiting compositionality and incremental class addition. To reduce supervision, test-time~\cite{feng2023diverse,yoon2024c,xiao2025dynaprompt,zhu2023prompt,sahir2024adaprompt} and continual adaptation strategies ~\cite{thengane2022clip,ding2022don,khan2023introducing, yu2024boosting, jha2024clap4clip, fu2025iap, zhou2025external, kang2025advancing, panos2025efficient} dynamically update CLIP. However, these approaches still couple adaptation with specific training or inference contexts and often suffer from degraded cross-task generalization.

A related line of work on generative personalization~\cite{gal2022image, ruiz2023dreambooth, kumari2023multi, zhang2023prospect, agarwal2025image, perera2025descriminative, wang2025multi} learns new tokens within CLIP’s vocabulary to represent novel or user-specific entities. However, these embeddings are optimized for image synthesis and fail to generalize to the broad range of downstream tasks where CLIP is applied. In contrast, we adapt CLIP’s text embeddings via subspace-guided objectives that balance semantic consistency and visual separability, yielding a unified, few-shot, plug-and-play framework for enhancing CLIP across diverse tasks~\cite{wang2022cris, luddecke2022image, esmaeilpour2022zero, yu2023convolutions, yan2023clip, lin2023clip, yu2023turning, cheng2024yolo}.

%% file: sec/3_methodology.tex
\section{Methodology}
\label{sec:method}

\begin{figure}[t]
\begin{center}
\includegraphics[width = 0.8\linewidth]{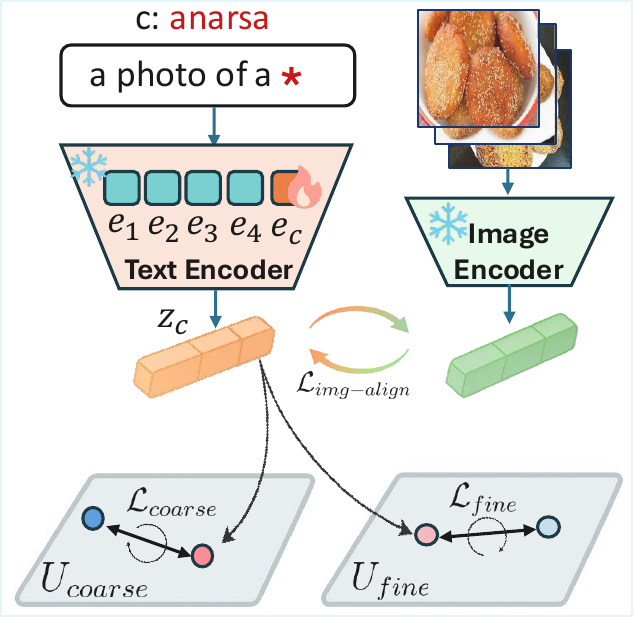}
\end{center}
\caption{
Overview of \modelname. Given a target class (e.g., \texttt{anarsa}) and a few reference images, the learnable token $e_c$ replaces the placeholder ``*'' in the prompt and is optimized while all CLIP encoders remain frozen. Subspace-Guided Optimization balances three objectives: image–text alignment ($\mathcal{L}_{\text{img-align}}$), coarse semantic anchoring ($\mathcal{L}_{\text{coarse}}$), and fine-grained separation ($\mathcal{L}_{\text{fine}}$), promoting discriminative yet semantically consistent embeddings.}
\label{fig:overview}
\end{figure}

\modelname, summarized in~\cref{fig:overview}, learns a class-specific text representation by optimizing only the embedding $e_c$ of the placeholder token ``$*$'' corresponding to a new class $c$, while keeping both CLIP encoders $f_{\text{img}}$ and $f_{\text{text}}$ fixed. Given an exemplar set $I_c = \{x_i\}_{i=1}^N$ corresponding to the class, the prompt $p_c$ (e.g., ``a photo of \(*\)'') is encoded into $z_c = f_{\text{text}}(p_c)$, initialized from the base CLIP text embedding $z_c^0 $. Adaptation is governed by three losses: a similarity-alignment loss encouraging high  $\operatorname{sim}\!\left(f_{\text{img}}\,(x_i),\; z_c\right)
$ over exemplars, and two additional loss terms that exploit the coarse-to-fine PCA decomposition of CLIP’s text space. These terms push $z_c$ to remain anchored to its broad semantic neighborhood along coarse components, while enforcing separation from visually confusable negatives 
$\mathcal{N}=\{n_k\}_{k=1}^K$ via their embeddings $z_{n_k}$, along fine components. We now motivate the design of our approach and delineate each component in detail.

\begin{figure*}[t]
\begin{center}
\includegraphics[width = 1.01\linewidth]{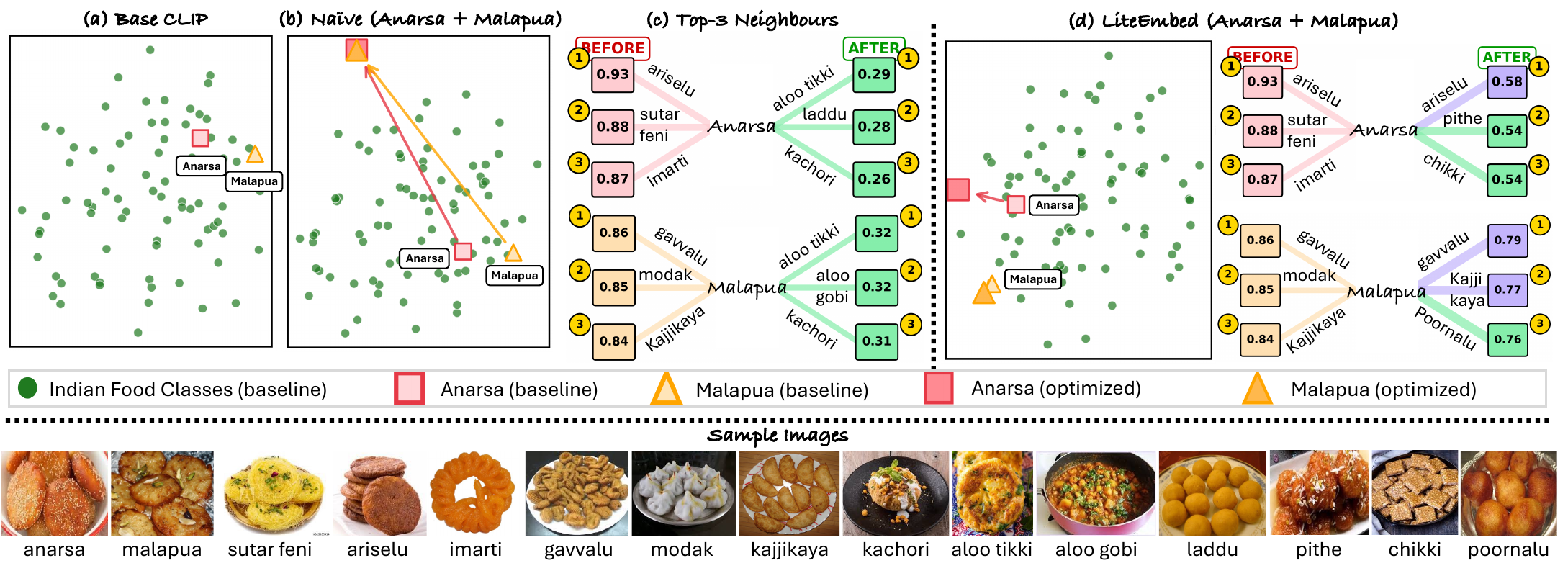}
\end{center}
\caption{
t-SNE visualization of CLIP’s text embeddings for \texttt{anarsa} and \texttt{malapua}. (a) Base CLIP embeddings before any adaptation. (b) Naïve image-only optimization leads to discriminative collapse, bringing the two classes too close. (c) Their nearest neighbors shift from related sweets to visually similar but semantically unrelated snacks, indicating semantic drift. (d) Our Subspace-Guided Optimization restores separation while preserving semantic relationships. Sample class images are shown below.
}

\label{fig:motivation_tsne_v2}
\end{figure*}

\subsection{Motivation}
\label{subsec:motivation}
We begin by examining how CLIP behaves for under-represented classes. Consider the example of \texttt{anarsa}, a traditional Indian sweet from the Indian Food Images dataset. Its base CLIP text embedding $z_c^0$ exhibits a very low average image-text similarity (0.19) and fails completely in classification (0\% accuracy), indicating poor alignment with the visual feature space (\cref{fig:motivation_tsne_v2} (a)).

To rectify this misalignment, we first refine the learnable token embedding $e_c$ for class $c$ (\texttt{anarsa}) to improve its alignment with the exemplar set $I_c$, leveraging image-text correspondence as CLIP’s principal supervisory signal. Formally, we optimize the alignment loss:

\[
\mathcal{L}_{\text{img-align}} = - \frac{1}{N} \sum_{i=1}^{N} \operatorname{sim}(f_{\text{img}}(x_i),\, f_{\text{text}}(p_c(e_c))),
\]  and update the embedding via gradient descent as $e_c \gets e_c - \eta \, \nabla_{e_c} \mathcal{L}_{\text{img-align}}$, where $\eta$ is the learning rate.

While such alignment-only optimization boosts performance for the adapted class, with \texttt{anarsa} achieving 100\% accuracy when evaluated for classification against baseline CLIP embeddings, it quickly exposes two critical issues when multiple classes are adapted sequentially:
\begin{enumerate}
    \item \textbf{Discriminative collapse}: When we subsequently optimize a new embedding (from scratch) for another visually similar sweet, \texttt{malapua}, the classification accuracy of \texttt{anarsa} drops sharply to 30\%. The t-SNE visualization in~\cref{fig:motivation_tsne_v2} (b) reveals that the optimized embeddings for both classes almost completely overlap, indicating that the optimization has captured shared visual cues (golden-brown circular texture) but failed to encode discriminative detail.

    \item \textbf{Semantic drift}: The optimized embedding for \texttt{anarsa} no longer resides within its original semantic neighborhood. In the baseline CLIP space, its nearest neighbors are related Indian sweets such as \texttt{ariselu}, \texttt{sutar feni}, and \texttt{imarti} (\cref{fig:motivation_tsne_v2}(c)). After naive optimization, these shift to visually similar but semantically unrelated snacks like \texttt{kachori} and \texttt{aloo tikki}, which are not sweets. Quantitatively, the maximum cosine similarity to any CLIP text embedding drops from about 0.93 to 0.29, confirming a substantial loss of semantic connectivity in the text manifold.
\end{enumerate}

This shows that alignment-only optimization, while visually faithful, distorts the embedding’s semantic neighborhood and fails to enhance discriminative power. An effective adaptation strategy must therefore preserve \textbf{global semantic alignment}, while avoiding \textbf{discriminative collapse} with related fine-grained classes. 

Accordingly, we pose a central question: \emph{can CLIP’s embedding space itself reveal how a novel class such as anarsa should be encoded, retaining its global semantics (e.g., Indian sweets) while enhancing separation from visually confusable classes like malapua?} This motivates our analysis of CLIP’s latent structure, presented next.

\subsection{Understanding CLIP’s Embedding Space}
\label{subsec:pca}

\begin{figure}[t]
\begin{center}
\includegraphics[width =\linewidth]{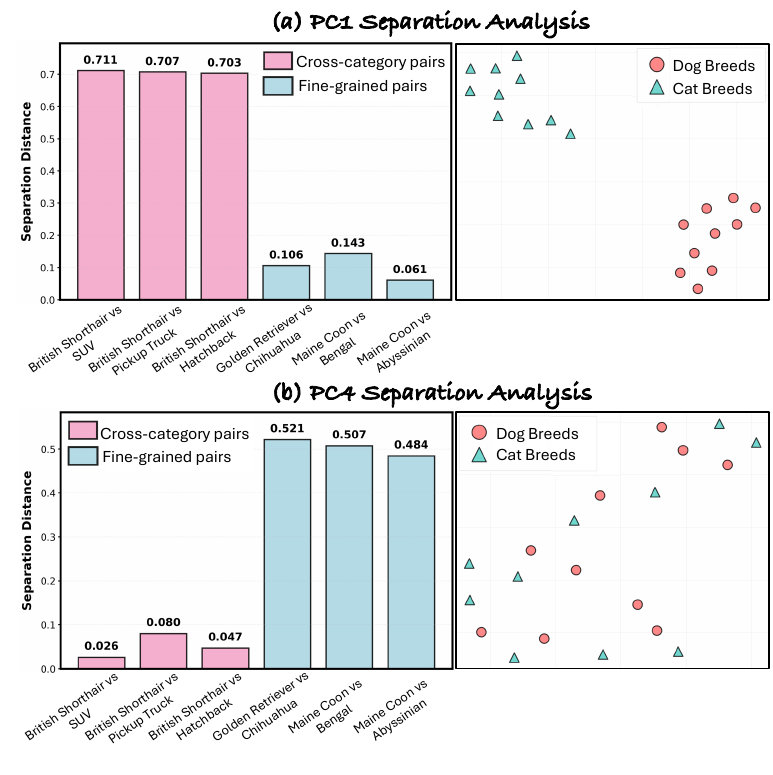}
\end{center}
\caption{PCA Analysis of CLIP's Text Embedding Space. 
}
\label{fig:pca_analysis}
\end{figure}

To examine the semantic structure of CLIP’s text embedding space, we conduct PCA over embeddings spanning diverse categories such as animals, vehicles, and food items. ~\cref{fig:pca_analysis} visualizes the separations along the first ($PC_1$) and fourth ($PC_4$) principal components. In (a), $PC_1$ captures coarse semantic variance: cross-category pairs (e.g., \texttt{British Shorthair} vs \texttt{SUV}) show markedly higher separation distances than fine-grained pairs (e.g., \texttt{Maine Coon} vs \texttt{Bengal}), indicating that high-variance directions encode coarse semantic structure. The accompanying t-SNE plot reveals two distinct clusters corresponding to dog and cat breeds, consistent with this coarse organization. In contrast, (b) shows $PC_4$, where fine-grained distinctions dominate and cross-category distances shrink, suggesting that lower-variance directions capture localized, discriminative semantics. The corresponding t-SNE projection exhibits mixed clusters with finer intra-breed separation, reinforcing this observation.

Overall, CLIP’s text space demonstrates a hierarchical organization: early components preserve broad semantic alignment, while later ones capture fine-grained variability, an insight that underpins our subspace-guided optimization framework for balanced alignment and discrimination.

\subsection{Subspace-Guided Optimization}

Given a new class $c$, our goal is to learn a text embedding $e_c$ that can be directly inserted into CLIP’s text encoder and used across downstream tasks. 

For each class, we define two reference neighborhoods: a \textbf{coarse set} $\mathcal{C}$, containing broader semantic groups suggested by an LLM that $c$ should stay aligned with (e.g., ``Indian sweet'', ``dessert'', ``snack''), and a \textbf{fine set} $\mathcal{F}$, containing visually similar yet distinct classes (e.g., \textit{malapua} for \textit{anarsa}). The fine set $\mathcal{F}$ constitutes the repository of hard-negative exemplars and is systematically constructed through a two-stage pipeline. In the first stage, a large language model (LLM) enumerates visually proximate candidates for each class. Subsequently, a CLIP-based filtering mechanism prunes this set by retaining candidates that exhibit the high similarity between their text embeddings and the image embeddings of the exemplars, thereby preserving only the most visually confusable instances.

We obtain CLIP text embeddings for all classes in these sets, each of dimensionality $d \in \{512, 768\}$ depending on the backbone. Performing PCA on the union $\mathcal{P} = \mathcal{C} \cup \mathcal{F}$ yields an orthogonal basis 
$U = [u_1, u_2, \dots, u_D]$, where $u_i \in \mathbb{R}^d$ and $D \leq d$. 
Guided by our PCA analysis in~\cref{fig:pca_analysis}, 
the top-variance components capture coarse semantics (broad category distinctions), while the low-variance ones encode fine-grained variations. We thus split $U$ into $U_{\text{coarse}} = [u_1, \dots, u_k]$ and $U_{\text{fine}} = [u_{k+1}, \dots, u_D]$, which serve as complementary subspaces for optimization.

The \textbf{coarse alignment loss} anchors $e_c$ to its semantic neighborhood along high-variance directions,
\[
\mathcal{L}_{\text{coarse}} 
= \frac{1}{|\mathcal{C}|} \sum_{p \in \mathcal{C}}
\big( 1 - \operatorname{sim}(U_{\text{coarse}}^\top z_c,\, U_{\text{coarse}}^\top z_p) \big),
\]
while the \textbf{fine separation loss} reduces similarity to visually confusable classes along low-variance directions,
\[
\mathcal{L}_{\text{fine}} 
= \frac{1}{|\mathcal{F}|} \sum_{n \in \mathcal{F}}
\operatorname{sim}(U_{\text{fine}}^\top z_c,\, U_{\text{fine}}^\top z_n).
\]
The final objective combines these with the standard image–text alignment loss:
\[
\mathcal{L} =
\mathcal{L}_{\text{img-align}} +
\lambda_1 \mathcal{L}_{\text{coarse}} +
\lambda_2 \mathcal{L}_{\text{fine}}.
\]

This formulation anchors $e_c$ to its coarse semantic neighborhood while maintaining discrimination against visually similar classes, enabling few-shot, training-free adaptation without retraining CLIP, preserving compositionality and plug-and-play use across downstream tasks.

\subsection{Putting It All Together}
When applied back to the motivating example of \texttt{anarsa} and \texttt{malapua} (\cref{subsec:motivation}), our subspace-guided optimization effectively resolves both issues observed with naïve alignment. As shown in~\cref{fig:motivation_tsne_v2} (d), the optimized embeddings for the two classes are now well-separated, preventing the overlap that previously caused \textbf{discriminative collapse}. At the same time, \textbf{semantic drift} is mitigated. Nearest-neighbor analysis in~\cref{fig:motivation_tsne_v2} (d) reveals that the optimized \texttt{malapua} embedding remains surrounded by other Indian sweets such as \texttt{gavvalu}, \texttt{poornalu}, and \texttt{kajjikaya}, many of which were its original neighbors in CLIP’s base embedding space. 

These results show that \modelname{} attains both fine-grained visual discriminability and coarse semantic fidelity: fine-subspace movement separates confusable classes, while coarse-subspace anchoring preserves linguistic and semantic grounding.

%% file: sec/4_results.tex
\section{Experiments}
\label{sec:results}

\textbf{Datasets:} We evaluate on 8 diverse classification datasets spanning both public and newly curated sources. The public ones include Indian Food Images~\cite{food2022indian}, Korean Celebrities~\cite{seo2023new}, and TV100~\cite{zhou2024tv100}. The remaining five namely Indian Singers, Indian Actors, Game Characters, Fashion Outfits, and Landmarks were newly collected using iCrawler as part of our proposed NOVA benchmark. For each of these, we first queried an LLM to identify classes that have emerged after 2023 within the respective domain, and then scraped 50 images per class. Please see supplementary material for the full class lists and scraping prompts. For downstream tasks, we use UECFood100~\cite{battini2023segmented} for segmentation and detection, Indian Food Images for retrieval, and CustomConcept101~\cite{kumari2023multi} for text-to-image generation.\\
\textbf{Experimental Setup:} We evaluate our method across three complementary settings: few-shot learning, continual learning, and downstream transfer. In the few-shot setup, we follow the standard N-shot classification protocol with $N \in \{1, 2, 4, 8, 16\}$ labeled samples per class. In the continual learning setup, we adopt a class-incremental scenario where classes appear sequentially, each task providing four samples from a single class; cumulative accuracy is reported over all classes observed so far and averaged across five random class orders to reduce ordering bias. For downstream transfer, we evaluate the learned embeddings within CLIP-based pipelines for text-guided retrieval, semantic segmentation, object detection, and text-to-image generation, without additional finetuning, to assess compatibility and reuse across tasks. All experiments use CLIP ViT-B/16 as the base model, except detection, where YOLO-World~\cite{cheng2024yolo} uses CLIP ViT-B/32 as its text encoder. Optimization is performed using Adam with a learning rate of $1 \times 10^{-4}$, a $1000$-step warm-up, and $5000$ total steps.

\textbf{Compared Approaches:}
We compare our method against zero-shot CLIP and a broad range of adaptation techniques, including prompt-learning methods (CoOp~\cite{zhou2022conditional}, CoCoOp~\cite{zhou2022learning}, MaPLe~\cite{khattak2023maple}), adapter-based approaches (CLIP-Adapter~\cite{gao2024clip}, TIP-Adapter-F~\cite{zhang2022tip}), and test-time adaptation methods (DiffTPT~\cite{feng2023diverse}, C-TPT~\cite{yoon2024c}, DynaPrompt~\cite{xiao2025dynaprompt}, PromptAlign~\cite{abdul2023align}). For continual learning, we evaluate Continual-CLIP~\cite{thengane2022clip}, ENGINE~\cite{zhou2025external}, MoE-Adapters4CL~\cite{yu2024boosting}, and an adapted CoOp variant that relearns context tokens after each task. For downstream transfer, we benchmark against CLIP-based task models, including CLIPSeg~\cite{luddecke2022image} for segmentation, YOLO-World~\cite{cheng2024yolo} for detection, Textual Inversion~\cite{gal2022image} for text-to-image generation, and standard CLIP for retrieval.\\

\subsection{Main Results}

By design, \modelname{} is trained in a continual manner, sequentially incorporating new classes without requiring joint access to earlier samples. However, for fair comparison with other methods, we categorize experiments into (i) few-shot classification, where all class samples are provided simultaneously, allowing joint adaptation, and (ii) continual learning, where classes are introduced sequentially.

We first assess \modelname{} under the standard few-shot classification setup with four samples per class provided simultaneously, and report the 4-shot accuracies in~\cref{tab:mainReults-fewshot}. \modelname{} decisively outperforms existing approaches across all datasets, with an average improvement of 14\% over the second-best approach (CoOp) and 35.6\% over baseline CLIP zero-shot. The most notable improvements occur on low-alignment domains such as Fashion Outfits (+37.8\%) and TV100 (+33.1\%), where CLIP’s zero-shot performance is especially weak. Extended \(1/2/8/16\)-shot results appear in the supplementary material.

\begin{figure*}[h]
\begin{center}
\includegraphics[width = 1.01\linewidth]{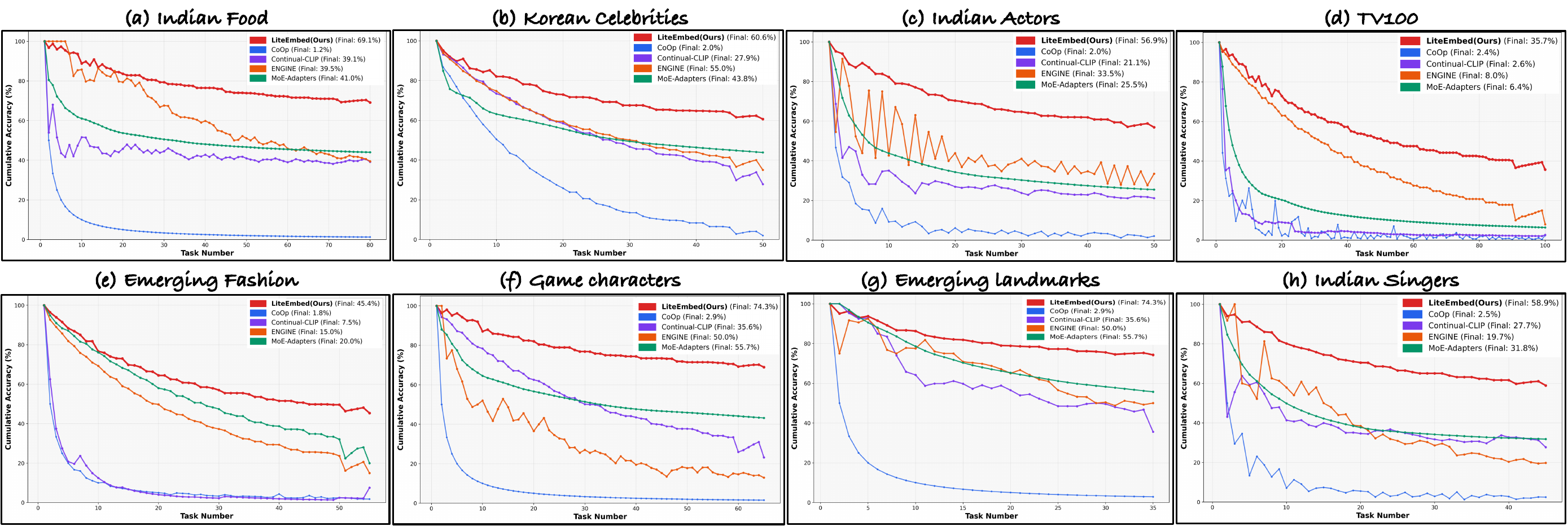}
\end{center}
\caption{Comparison of 4-shot continual adaptation results across sequentially added classes. \modelname{} in red, ENGINE in orange, CoOp in blue, Continual-CLIP in purple, and MoE-Adapters in green; best viewed in color.}
\label{fig:continualExpsPlot}
\end{figure*}

We next evaluate \modelname{} in its native continual setup, shown in~\cref{fig:continualExpsPlot}, where adaptation proceeds sequentially over new class groups. \modelname{} consistently outperforms baselines, achieving 32\% average higher final accuracy over the second-best approach (ENGINE) and significantly lower forgetting, whereas CoOp shows near-complete forgetting of previously learned context tokens. Minor performance drops across tasks, primarily arise from increased candidate classes (rather than forgetting), highlighting \modelname’s ability to integrate classes progressively while maintaining stable performance.
 
\begin{table*}[ht]
\centering
\resizebox{\linewidth}{!}{%
\begin{tabular}{lccccccccc}
\toprule
\textbf{Method} & \textbf{Indian Food} & \textbf{Korean Celebrities} & \textbf{Indian Singers} & \textbf{Indian Actors} & \textbf{Game Characters} & \textbf{Fashion Outfits} & \textbf{TV100} & \textbf{Landmarks} & \textbf{Average} \\
& (80 classes) & (50 classes) & (50 classes) & (50 classes) & (65 classes) & (55 classes) & (60 classes) & (35 classes) & \\
\midrule
\textbf{CLIP Zero-Shot} & 39.1 & 27.9 & 27.7 & 21.6 & 23.2 & 7.6 & 2.6 & 35.6 & 23.2 \\
\midrule
\multicolumn{10}{l}{\textit{Prompt learning methods}} \\
CoOp~\cite{zhou2022conditional}~\tabpub{CVPR'22} & \underline{53.3} & \underline{52.6} & \underline{55.2} & \underline{43.2} & \underline{56.5} & \underline{30.7} & 1.1 & 58.2 & 44.0 \\
CoCoOp~\cite{zhou2022learning}~\tabpub{IJCV'22} & 46.9 & 44.5 & 37.2 & 26.5 & 54.9 & 12.3 & 1.2 & 57.1 & 35.1 \\
MaPLe~\cite{khattak2023maple}~\tabpub{CVPR'23} & 50.9 & 50.7 & 32.8 & 34.8 & 55.2 & 17.5 & 0.9 & \underline{65.7} & 38.6 \\
\midrule
\multicolumn{10}{l}{\textit{Adapter-based methods}} \\
CLIP-Adapter~\cite{gao2024clip}~\tabpub{IJCV'24} & 43.8 & 37.5 & 30.6 & 21.2 & 36.0 & 11.4 & 0.7 & 49.3 & 28.9 \\
TIP-Adapter-F~\cite{zhang2022tip}~\tabpub{ECCV'22} & 52.3 & 40.2 & 39.0 & 33.0 & 40.8 & 11.1 & \underline{8.5} & 57.9 & 35.4 \\
\midrule
\multicolumn{10}{l}{\textit{Test-time adaptation methods}} \\
DiffTPT~\cite{feng2023diverse}~\tabpub{ICCV'23} & 42.3 & 27.6 & 20.6 & 17.7 & 23.6 & 8.1 & 1.7 & 38.6 & 22.4 \\
PromptAlign~\cite{abdul2023align}~\tabpub{NeurIPS'23} & 23.8 & 26.2 & 17.8 & 16.1 & 23.4 & 6.1 & 2.4 & 33.9 & 18.7 \\
C-TPT~\cite{yoon2024c}~\tabpub{ICLR'24} & 39.2 & 23.1 & 19.1 & 15.9 & 21.9 & 6.6 & 1.0 & 36.1 & 20.5 \\
DynaPrompt~\cite{xiao2025dynaprompt}~\tabpub{ICLR'25} & 41.6 & 24.0 & 19.6 & 21.0 & 30.4 & 6.8 & 2.6 & 45.7 & 24.0 \\
\midrule
\textbf{~\modelname~(Ours)} & \textbf{69.1} & \textbf{60.6} & \textbf{58.9} & \textbf{56.9} & \textbf{68.9} & \textbf{45.4} & \textbf{35.7} & \textbf{74.3} & \textbf{58.7} \\
\bottomrule
\end{tabular}%
}
\caption{Few-Shot Classification Accuracy Results (4-shot).~\modelname{} consistently outperforms all baselines across 8 diverse datasets.}
\label{tab:mainReults-fewshot}
\end{table*}

\begin{figure*}[h]
\begin{center}
\includegraphics[width = 0.9\linewidth]{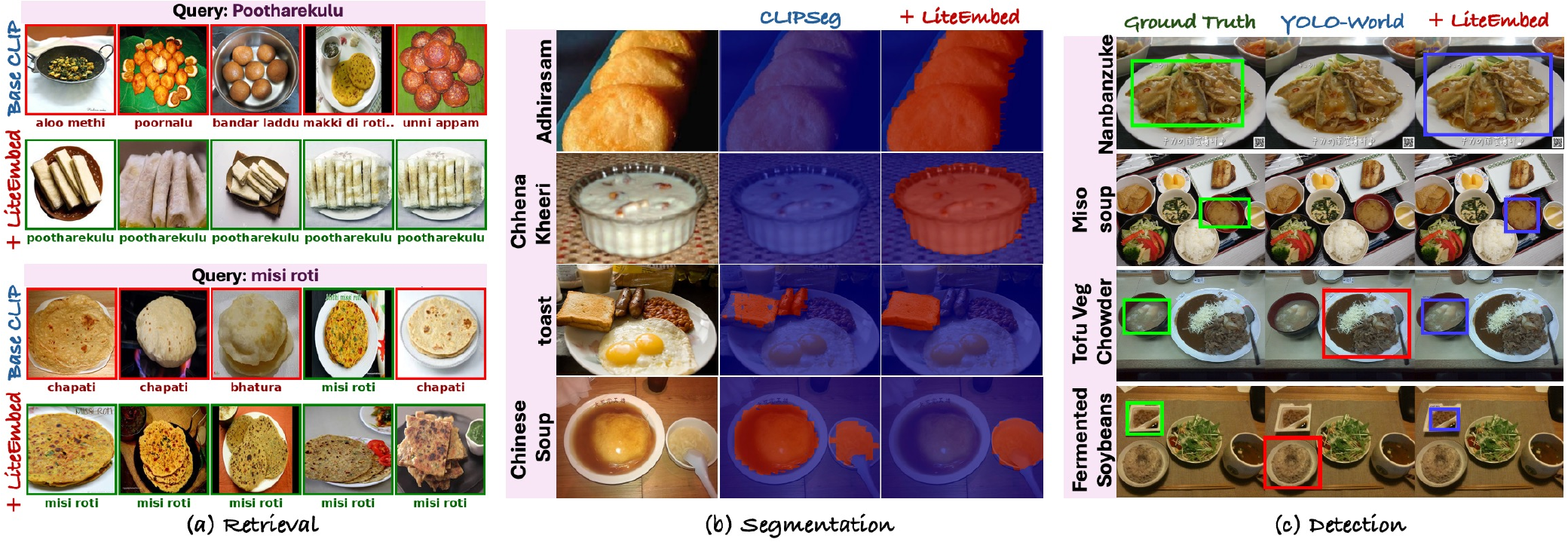}
\end{center}
\caption{Qualitative results across downstream tasks. \modelname\ consistently improves retrieval, segmentation, and detection quality.}
\label{fig:downstreamCombined_RSD}
\end{figure*}

\begin{figure}[h]
\begin{center}
\includegraphics[width = 0.9\linewidth]{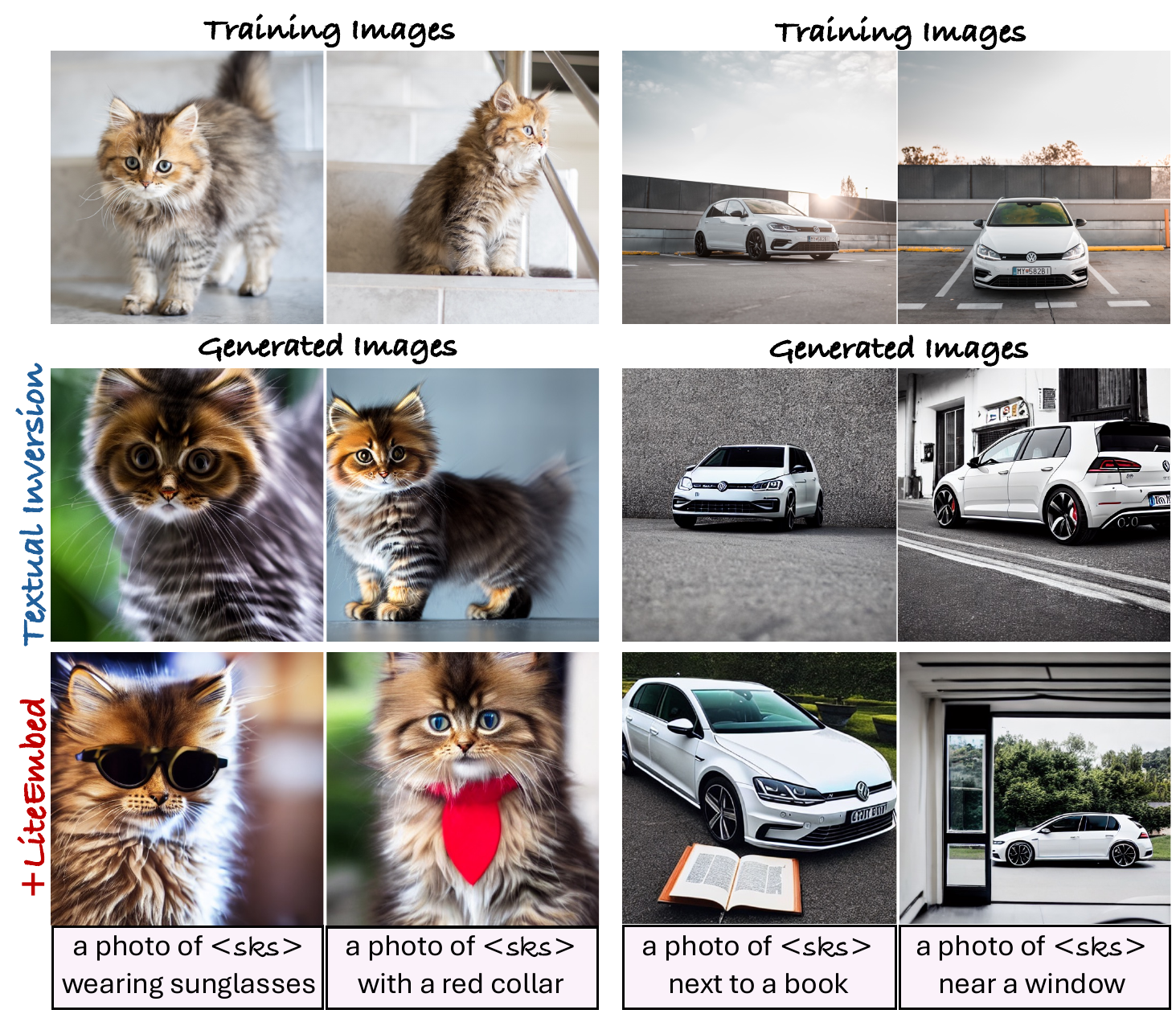}
\end{center}
\caption{Textual Inversion results showing that \modelname\ enables more editable image generation compared to baseline TI. Here, $<sks>$ in prompts denotes the learned class embedding.}
\label{fig:textual_inversion}
\end{figure}

\subsection{Evaluation on Downstream Tasks}
To assess the generality of \modelname{} beyond few-shot classification, we evaluate it on downstream tasks that rely on CLIP text embeddings, including retrieval, segmentation, detection, and text-to-image generation. Unlike prior methods optimized solely for classification, \modelname{} requires no task-specific retraining; the learned embedding simply replaces the original CLIP text embedding wherever the text encoder is used. This illustrates its plug-and-play nature and broad applicability across vision–language tasks. We benchmark against task-specific CLIP-based models, including CLIPSeg for segmentation, YOLO-World for detection, and Textual Inversion for generative adaptation.

\textbf{Retrieval:} We evaluate~\modelname{} on text-based image retrieval using the Indian Food Images dataset. Replacing the original CLIP text embeddings with~\modelname{} embeddings, we observe substantial gains in retrieval precision, with Precision@5 improving from 26.2\% to 68.7\% across 80 food classes. For the 20 most underrepresented dishes, the gains are even more pronounced, with Precision@5 improving from 13.3\% to 93.6\%. Qualitatively,~\cref{fig:downstreamCombined_RSD} (a) illustrates common failure modes of baseline CLIP: for the class \texttt{Pootharekulu}, CLIP retrieves images from unrelated categories, while for \texttt{Misi Roti}, it confuses visually similar breads such as \texttt{Bhatura} and \texttt{Chapati}. In contrast,~\modelname{} embeddings retrieve accurate images across the top-5 positions, highlighting their ability to capture distinctive visual characteristics for rare classes.

\textbf{Segmentation:} CLIPSeg consumes a text prompt and outputs a corresponding segmentation mask. On the Indian Food Dataset (IFD), which contains many underrepresented categories, baseline CLIPSeg fails to generate masks for 23 of the 80 classes (zero-coverage classes, where coverage denotes the percentage of segmented pixels), underscoring its limitations on rare classes. Although IFD lacks ground-truth masks and precise quantitative evaluation is not feasible, we observe that \modelname{} produces nonzero segmentation coverage for all classes after adaptation, eliminating all zero-coverage cases. Qualitative results in~\cref{fig:downstreamCombined_RSD} illustrate this effect. Panel (b), first row, shows that baseline CLIPSeg fails to segment \texttt{adhirasam}, a failure that is fully recovered with \modelname{}.

UECFOOD100 provides ground-truth segmentation masks and includes multi-object images, where CLIPSeg often lacks discrimination and segments incorrect regions. Quantitatively, \modelname{} delivers a substantial IoU improvement, increasing overall IoU from $0.43$ to $0.71$. Qualitative examples in~\cref{fig:downstreamCombined_RSD}(b) illustrate this effect: in the third row, baseline CLIPSeg segments additional objects beyond the intended \texttt{toast}, whereas \modelname{} correctly isolates the target; the fourth-row \texttt{Chinese soup} example shows a similar correction. These results confirm that \modelname{} significantly enhances both coverage and mask fidelity for rare and visually confusable categories.

\textbf{Detection:} We evaluate~\modelname{} on UECFOOD100 open-vocabulary detection using YOLO-World, replacing class text embeddings with the optimized embeddings. We report mean IoU (mIoU) as the primary evaluation metric. With~\modelname{}, overall mIoU notably increases from 0.37 to 0.72, compared to using baseline CLIP embeddings. Qualitatively,~\cref{fig:downstreamCombined_RSD} (c) illustrates typical failure modes of YOLO-World with baseline embeddings: in the first two rows, objects of interest (\texttt{Nanbanzuke} and \texttt{Miso soup}) are entirely missed, while in others, incorrect items are instead detected. Using~\modelname{} embeddings corrects these errors, enabling accurate detection of all target objects and improving localization precision. 

\textbf{Generation:}
We extend~\modelname{} to custom image generation via Textual Inversion (TI), which learns new class tokens by optimizing embeddings in CLIP’s text space. We replace ~\modelname’s alignment loss $\mathcal{L}_{align}$ with TI’s reconstruction loss, keeping all other aspects of training unchanged. This ensures the learned embedding generates images faithful to the class. Qualitative results~\cref{fig:textual_inversion} on examples from the CustomConcept101 dataset show that while baseline TI struggles with compositional prompts,~\modelname{} learned embeddings successfully generate images such as a custom cat wearing sunglasses, demonstrating improved editability without sacrificing reconstruction fidelity. Additional results are presented in the supplementary material.

\subsection{Ablations}
Ablation experiments on the Indian Food Images dataset evaluate the contribution of each loss component in our optimization objective.~\cref{tab:ablation} reports classification accuracy and the average image–text cosine similarity between class images and their corresponding ground-truth text embeddings. Introducing the image–text alignment loss $\mathcal{L}_{\text{img-align}}$ improves alignment scores over base CLIP but does not substantially boost accuracy, indicating limited generalization when optimized alone. Adding the coarse alignment loss $\mathcal{L}_{\text{coarse}}$ leads to modest accuracy gains, likely due to reduced confusion between classes belonging to different semantic neighborhoods, while also preserving compositional structure within CLIP’s text space. Finally, incorporating the fine separation loss $\mathcal{L}_{\text{fine}}$ enhances discriminative power, yielding the highest overall accuracy. These results collectively demonstrate that the full subspace-guided objective achieves a balanced trade-off between preserving semantic consistency and enhancing visual discrimination, leading to the highest overall performance. We further provide ablations across different CLIP backbones and experiments analyzing the choice of $k$ for subspace-guided optimization in the supplementary material.

\begin{table}[t]
\centering
\caption{
Ablation study on the Indian Food Images dataset.}
\vspace{0.5em}
\resizebox{\linewidth}{!}{
\begin{tabular}{lcc}
\toprule
\textbf{Method} & \textbf{Accuracy (\%)} & \textbf{Avg. Image–Text Cosine Similarity} \\
\midrule
Base CLIP & 39.1 & 0.28 (range: 0.18-0.35) \\
+ $\mathcal{L}_{\text{img-align}}$ & 45.3 & 0.65 (range: 0.56-0.71)  \\
+ $\mathcal{L}_{\text{img-align}}$ + $\mathcal{L}_{\text{coarse}}$ & 52.6 & 0.48 (range: 0.40-0.53) \\
+ $\mathcal{L}_{\text{img-align}}$ + $\mathcal{L}_{\text{coarse}}$ + $\mathcal{L}_{\text{fine}}$ & \textbf{69.1} & 0.39 (range: 0.33-0.42) \\
\bottomrule
\end{tabular}
}
\label{tab:ablation}
\end{table}

%% file: sec/5_conclusion.tex
\section{Conclusion}

In this work, we addressed the challenge of adapting CLIP to rare and culturally diverse concepts, where standard text embeddings often fail to align with visual semantics. We introduced~\modelname{}, a plug-and-play framework that refines CLIP’s text embeddings through Subspace-Guided Optimization using only a few reference images. Our formulation balances coarse semantic anchoring with fine-grained discriminative adaptation, yielding embeddings that integrate seamlessly into CLIP-based models across tasks such as retrieval, segmentation, detection, and generation, all without task-specific retraining. Extensive experiments demonstrate consistent improvements over prior adaptation methods and validate the generality of our approach. Future work will investigate applying subspace-guided adaptation to other vision–language models, studying its generality across diverse multimodal backbones.

%% file: sec/X_suppl.tex

\appendix
\section{Appendix}
\label{sec:appendix}

In Section~\ref{subsec:trendsShots}, we present extended 1/2/4/8/16-shot results, highlighting how \modelname{} scales with data availability and where gaps between our method and existing baselines are most pronounced. In Section~\ref{subsec:dset}, we provide details of our data collection pipeline and the LLM prompts used to construct the NOVA benchmark. In Section~\ref{subsec:ablations}, we show additional results across CLIP backbones to demonstrate the architecture-agnostic nature of our approach. In Section~\ref{subsec:tiVSclip}, we compare against Textual Inversion to illustrate the limitations of generation-oriented personalization when applied to large-scale discriminative settings. In Section~\ref{subsec:imageRef}, we further analyze a reference-image baseline to clarify why operating purely in image space undermines downstream compositionality. Finally, we include expanded qualitative examples across retrieval, segmentation, detection, and generation tasks in Section~\ref{subsec:qualSupp} as well as additional analysis of our PCA-based subspace choice in Section~\ref{subsec:kChoice}.

\subsection{1/2/8/16 few shot results}
\label{subsec:trendsShots}
\begin{figure*}[h]
\begin{center}
\includegraphics[width = 1.01\linewidth]{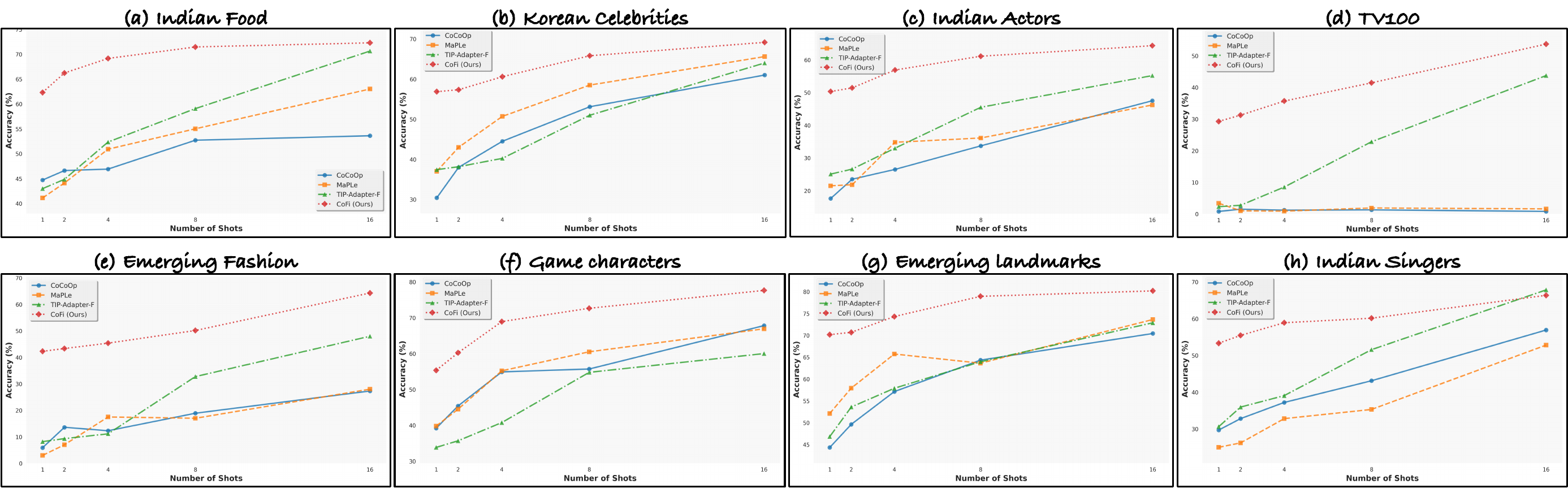}
\end{center}
\caption{Comparison of few-shot classification accuracy across 1-, 2-, 4-, 8-, and 16-shot settings on multiple datasets.}
\label{fig:trendsShots}
\end{figure*}

To further study how ~\modelname{} scales with data availability, we compare its performance against CoOp~\cite{zhou2022conditional}, MaPLe~\cite{khattak2023maple}, and TIP-Adapter-FTIP-Adapter-F~\cite{zhang2022tip} under 1-, 2-, 4-, 8-, and 16-shot settings (shown in Figure~\ref{fig:trendsShots}). \modelname{} consistently outperforms these baselines even in the 1- and 2-shot regimes, where the gap is especially pronounced. On datasets such as Indian Food, the improvement over prior methods reaches 15–20\% in the extreme low-shot setting, and on Indian Actors, the margin widens further to 25–35\%, highlighting \modelname{}’s strong generalization under minimal supervision. While TIP-Adapter-F begins to close the gap as more samples become available, it still trails \modelname{} across all shot counts. Moreover, existing approaches remain tied to fixed few-shot setups and do not naturally extend to incremental integration of new concepts. 

\subsection{Dataset scraping details}
\label{subsec:dset}
The five datasets in our NOVA benchmark, namely \textit{Indian Singers}, \textit{Indian Actors}, \textit{Game Characters}, \textit{Fashion Outfits}, and \textit{Landmarks}, were newly collected to evaluate CLIP's performance on post-2023 emerging content and culturally-specific domains. We developed a systematic data collection pipeline using iCrawler~\cite{icrawler} with Bing as the image source, chosen for its reliability and comprehensive content coverage. To ensure our benchmark targets content outside CLIP's pretraining distribution (which has a knowledge cutoff around 2021), we generated class lists using the following prompt with a large language model~\cite{brown2020language}:

\begin{figure}[h]
\centering
\noindent\begin{minipage}{0.45\textwidth}
\mdfsetup{%
middlelinewidth=1pt,
backgroundcolor=cyan!10,
innerleftmargin=0.5cm,
innerrightmargin=0.5cm,
roundcorner=15pt}
\begin{mdframed}
\vspace{0.02em}

Generate a list of 50 domain-specific classes that emerged or gained significant prominence after 2023, for each of Indian Singers, Indian Actors, Game Characters, Fashion Outfits, and Landmarks. Focus on entities that:
\begin{enumerate}
    \item have substantial visual presence online,
    \item represent diverse subcategories within the domain,
    \item are culturally specific or region-specific where applicable,
    \item are unlikely to have been well-represented in pre-2022 vision-language training datasets.
\end{enumerate}
For Indian Singers and Actors, prioritize emerging talents from recent films and music releases. For Game Characters, focus on notable characters popular games released in 2023-2024. For Fashion Outfits, select outfits from emerging fashion brands and designers that launched or gained international recognition post-2023. For Landmarks, include architectural structures and public installations completed after 2023.

\end{mdframed}
\end{minipage}
\label{fig:gpt_dset_classes_prompt}
\vspace{0.02em}
\end{figure}

\begin{figure*}[h]
\begin{center}
\includegraphics[width = 1.01\linewidth]{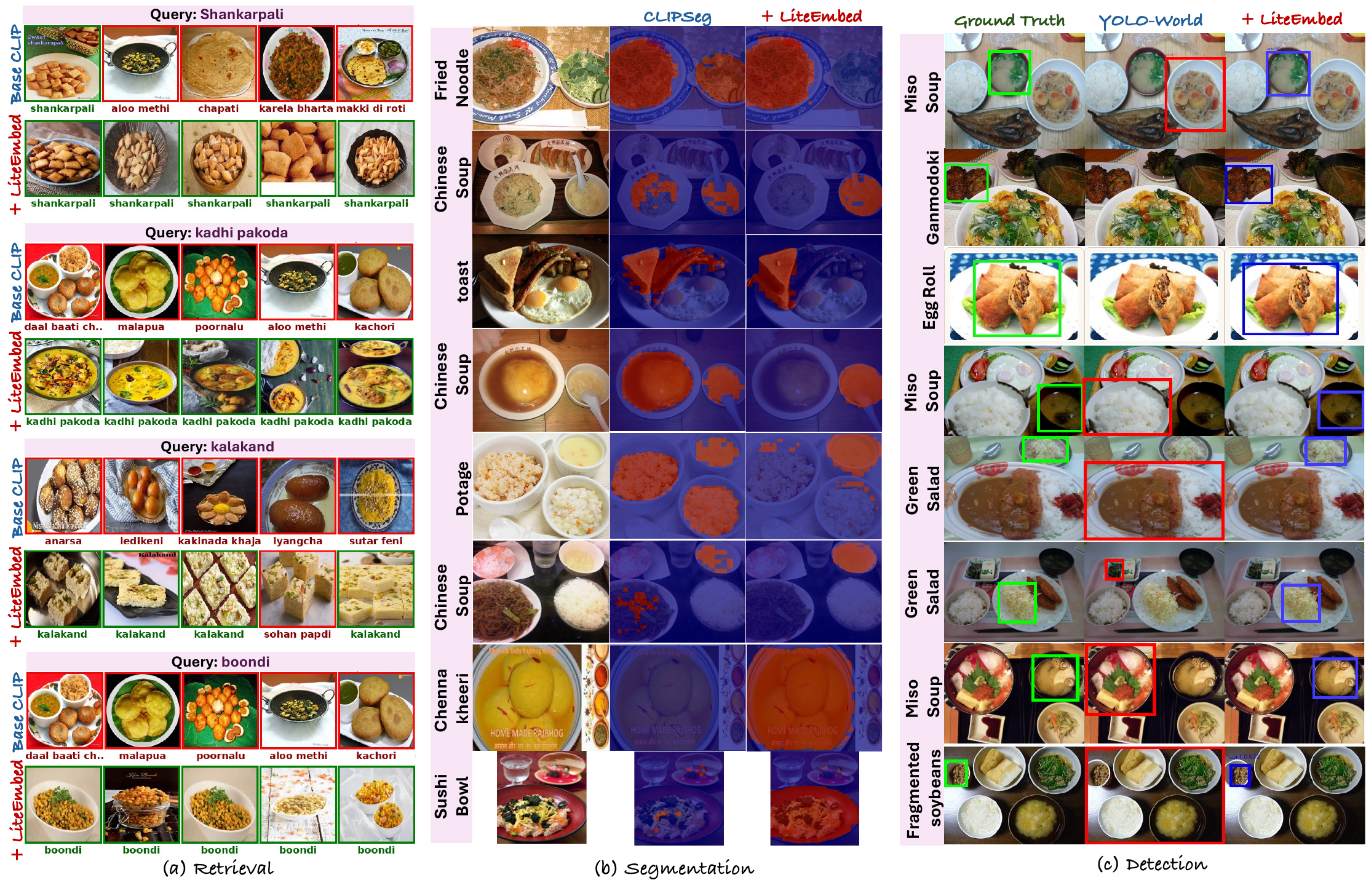}
\end{center}
\caption{Additional Qualitative results across downstream tasks. \modelname\ consistently improves retrieval, segmentation, and detection quality over baseline CLIP embeddings.}
\label{fig:qualRSDsupp}
\end{figure*}

\subsection{Ablation across CLIP backbones}
\label{subsec:ablations}
To demonstrate the generalizability of our approach across different model scales,
we evaluate on both ViT-B/32 and ViT-L/14 CLIP backbones (Table~\ref{tab:classification_baseline_vs_ours}).
Despite their substantial differences in model size and
patch resolution (32$\times$32 vs. 14$\times$14), LiteEmbed achieves consistent
improvements across both architectures: +33.3\% for ViT-B/32 
and +35.75\% for ViT-L/14, averaged over all the datasets.
This backbone-agnostic behavior demonstrates that our method inherits the architecture's
flexibility without requiring model-specific tuning.

\begin{table*}[h]
\centering
\caption{Classification Performance (Top-1 Accuracy \%): Baseline CLIP vs Our Textual Inversion Method}
\label{tab:classification_baseline_vs_ours}
\begin{tabular}{l|cc|cc}
\hline
\multirow{2}{*}{\textbf{Dataset}} & \multicolumn{2}{c|}{\textbf{ViT-B/32}} & \multicolumn{2}{c}{\textbf{ViT-L/14}} \\
\cline{2-5}
 & Baseline CLIP & LiteEmbed & Baseline CLIP & LiteEmbed \\
\hline

Indian Food          &  36.05 &  63.22  &  44.21 &  71.28     \\
Game Characters      &  25.31 &  64.13  &  31.15 &  69.21      \\
Indian Actors        &  18.91 &  51.12  &  21.80 &  59.34      \\
Indian Singers       &  23.46 &  55.67 &  31.85 &   60.23     \\
Landmarks     &  37.38 &  74.12   &  47.81 &  89.18      \\
TV100                &   2.38 &   34.11  &   3.51 & 38.13       \\
Fashion Outfits    &   7.68 &  42.38  &   9.31 & 51.16       \\
Korean Celebrities   &   25.23  &  54.61      & 27.16      &   64.43     \\
\hline
\textbf{Average}     & \textbf{21.59} & \textbf{54.92}       & \textbf{27.12} &   \textbf{62.87}     \\
\hline
\end{tabular}
\end{table*}
\subsection{Textual Inversion for classification}
\label{subsec:tiVSclip}
\begin{figure}[h]
\begin{center}
\includegraphics[width = 1.01\linewidth]{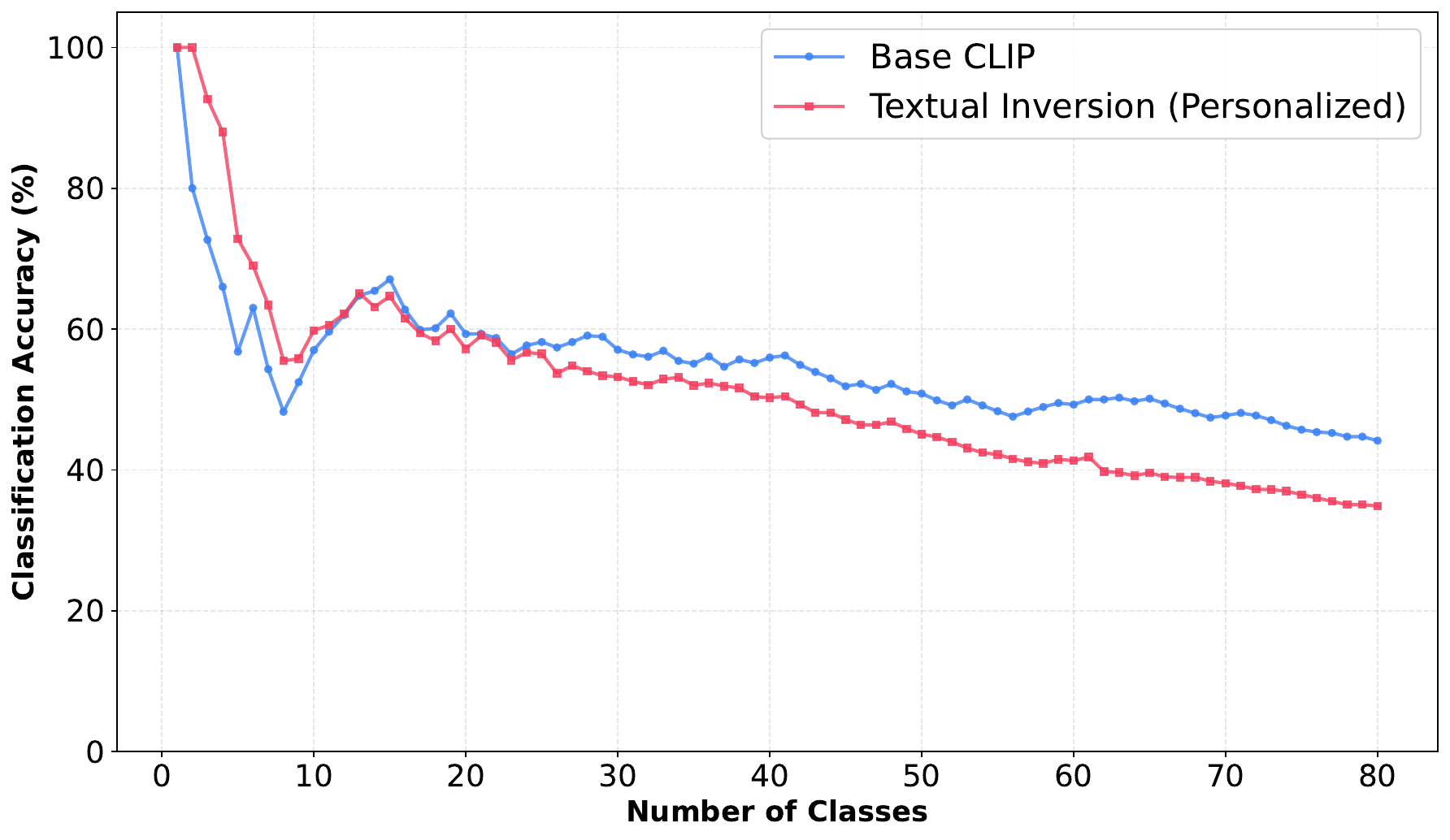}
\end{center}
\caption{Textual Inversion for classification.}
\label{fig:tiVSclip}
\end{figure}

While personalization methods like Textual Inversion~\cite{gal2022image} have demonstrated remarkable success in embedding new visual concepts directly into CLIP's vocabulary for generation tasks, their efficacy in downstream discriminative tasks remains underexplored. To investigate this, we conduct an incremental classification experiment on the Indian Food dataset (80 classes), comparing the zero-shot classification performance of base CLIP ViT-L/14 text embeddings against learned TI embeddings. Our experimental protocol follows an incremental learning setup: for each task k $\in$ [1, 80], we evaluate classification accuracy on k classes using both base CLIP embeddings (``a photo of [class]") and personalized TI embeddings (``a photo of $<$class$>$") as text encoders. As shown in Figure~\ref{fig:tiVSclip}, we observe that TI embeddings provide substantial gains (15-22\% improvement) for small vocabularies (8-10 classes), but this advantage rapidly diminishes as the number of classes grows. Beyond approximately 13 classes, base CLIP consistently outperforms TI, with the performance gap widening to 9.3\% at the full 80-class setting (44.2\% vs 34.9\% accuracy). 

This trend suggests that while TI's generation-optimized embeddings capture fine-grained visual details essential for reconstruction, they lack the semantic structure and inter-class relationships inherent in CLIP's pre-trained representations, which prove critical for robust discrimination across large vocabularies. These findings highlight a fundamental trade-off: embeddings optimized for generation (via diffusion reconstruction loss) may not transfer effectively to discriminative tasks, suggesting the need for multi-objective training strategies that balance both generative fidelity and discriminative power.

\subsection{Using reference images directly}
\label{subsec:imageRef}

We evaluate a few-shot reference-based baseline that bypasses CLIP's text encoder entirely by directly using mean image embeddings from four reference images as class representatives. As shown in Table~\ref{tab:imageRef}, this few-shot approach yields only a modest average gain of +4.04\% across seven diverse datasets and behaves inconsistently, even reducing performance relative to base CLIP on three of the seven datasets (Indian Food, Indian Singers, and Landmarks). Most of the apparent improvements occur on datasets where CLIP's original text embeddings perform extremely poorly (e.g., 3.49\% on TV100, 7.68\% on emerging fashion), suggesting that while this method can offer small boosts when text-based representations fail, it cannot meaningfully improve performance beyond that narrow regime.

In contrast, retrieval benefits more reliably, since image-to-image similarity is often stronger than text-to-image alignment for visual search. However, this baseline fundamentally breaks CLIP's vision–language alignment by operating entirely in image space, making it incompatible with downstream tasks that rely on text–image compositionality, such as open-vocabulary detection, referring segmentation, and text-to-image generation. It also cannot take advantage of linguistic compositionality (e.g., combining ``red" and ``car") or zero-shot generalization to unseen text prompts—both central to CLIP's utility.

By comparison, LiteEmbed learns embeddings directly within CLIP's native text space, preserving full compositionality and compatibility with the text encoder while providing consistent improvements across tasks.

\begin{table*}[h]
\centering
\caption{Classification and Retrieval Performance using ViT-B/16 with 4-shot Reference Image Embeddings}
\label{tab:imageRef}
\begin{tabular}{l|cc|cc}
\hline
\multirow{2}{*}{\textbf{Dataset}} & \multicolumn{2}{c|}{\textbf{Classification (Top-1 Acc \%)}} & \multicolumn{2}{c}{\textbf{Retrieval (P@5 \%)}} \\
\cline{2-5}
 & Baseline & Few-shot & Baseline & Few-shot \\
\hline

Indian Food          &  39.10 &  26.86 &  52.00 &  58.75 \\
Game Characters      &  29.86 &  42.76 &  49.85 &  59.38 \\
Indian Actors        &  23.06 &  34.48 &  27.20 &  51.60 \\
Indian Singers       &  31.13 &  28.79 &  39.11 &  42.00 \\
Landmarks     &  45.31 &  42.10 &  60.57 &  69.71 \\
TV100                &   3.49 &  17.51 &   6.60 &  35.80 \\
Fashion Outfits     &   7.68 &  16.29 &  30.91 &  45.82 \\
\hline
\end{tabular}
\end{table*}

\subsection{Additional qualitative results for all tasks}
\label{subsec:qualSupp}
\begin{figure}[h]
\begin{center}
\includegraphics[width = 1.01\linewidth]{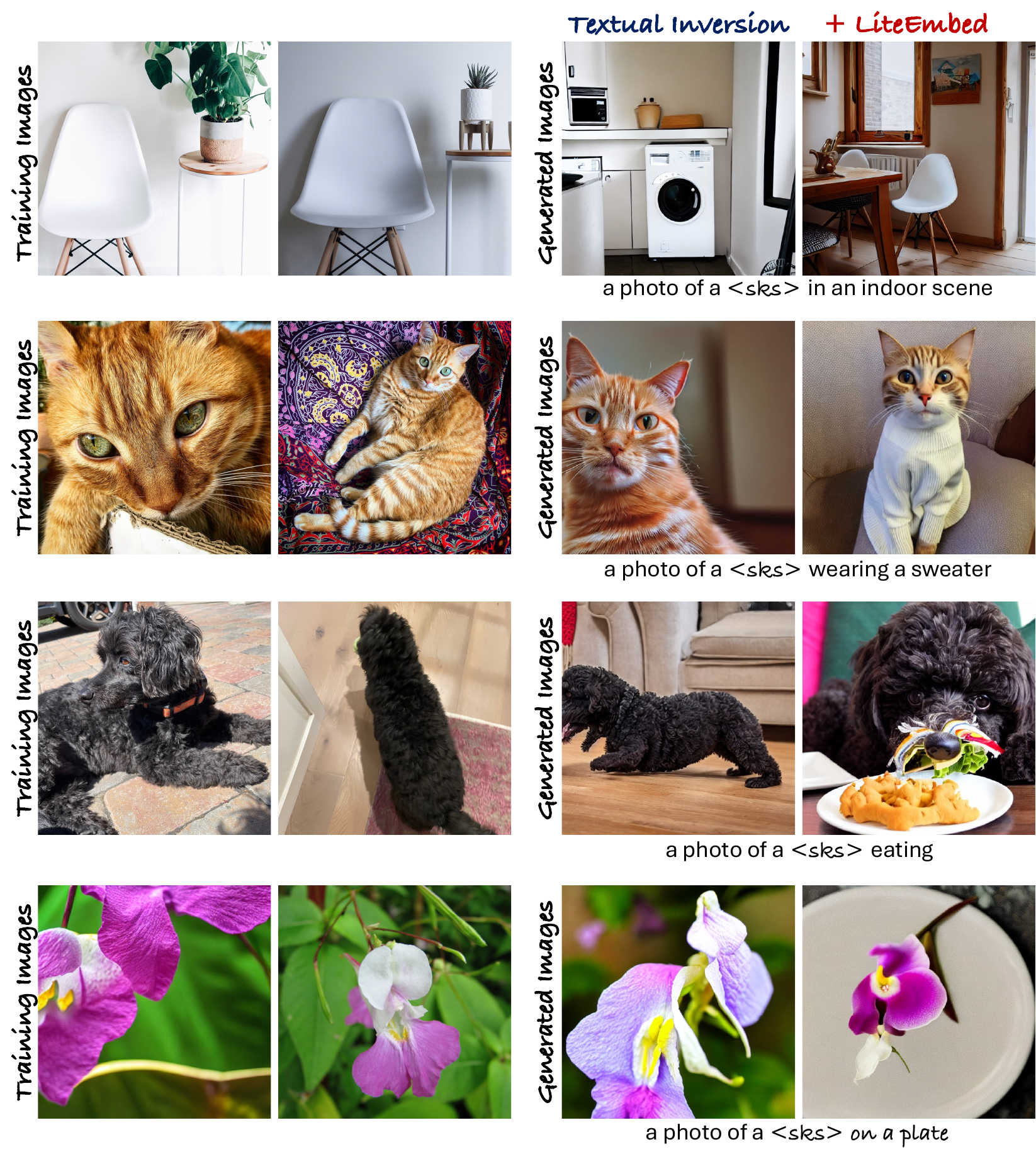}
\end{center}
\caption{Additional Qualitative results across downstream tasks. \modelname\ consistently improves retrieval, segmentation, and detection quality over baseline CLIP embeddings.}
\label{fig:genDTsupp}
\end{figure}

We show additional results in Figure~\ref{fig:genDTsupp} and~\ref{fig:qualRSDsupp} demonstrating LiteEmbed's strong adaptability to downstream tasks like retrieval, segmentation, detection and generation. 

\subsection{Choice of k (split) in PCA}
\label{subsec:kChoice}
We set $k \geq 2$ (i.e., we drop the first principal component) for the fine subspace. To justify this choice, we ran a PCA analysis on 75 classes drawn from five broad categories--dogs, cats, vehicles, food, and buildings. For each principal component, we quantified how much it favors coarse-grained (across categories) versus fine-grained (within a category) separation by taking the ratio of average cross-category distances to within-category distances.

PC1 shows a ratio of 5.02, meaning it separates classes across different categories about 5$\times$ more strongly than it separates classes within the same category. By contrast, PCs~2--5 have an average ratio of 1.62, suggesting they are far less biased toward coarse distinctions.

We also inspected the ten class pairs that PC1 separates the most, and all of them turned out to be cross-category (for example, \emph{cat vs.\ building}). When we evaluated the same pairs on PCs~2--5, their separation dropped by 76\%, meaning that without PC1, those pairs lose most of their discriminability. The reverse pattern also holds: within-category pairs that are highly separated on PC2 retain only 22\% of that separation when projected onto PC1.

Taken together, these results show a clear asymmetry: PC1 mainly captures broad, category-level differences, whereas PC2 and later components encode the fine-grained variations that matter for personalized classification. This is why we exclude PC1 and rely on PCs $k \ge 2$ for our task.
